%% file: main.tex
\documentclass[10pt,twocolumn,letterpaper]{article}

\usepackage{cvpr}              % To produce the CAMERA-READY version
\input{preamble}

\definecolor{cvprblue}{rgb}{0.21,0.49,0.74}
\usepackage[pagebackref,breaklinks,colorlinks,allcolors=cvprblue]{hyperref}
\usepackage{multirow}
\usepackage{bbding}
\usepackage{balance}
\def\paperID{12862} % *** Enter the Paper ID here
\def\confName{CVPR}
\def\confYear{2026}

\title{Multigrain-aware Semantic Prototype Scanning and Tri-Token Prompt Learning Embraced High-Order RWKV for Pan-Sharpening}

\author{
    Junfeng Li$^{1,2,\dagger}$, Wenyang Zhou$^{3,\dagger}$, Xueheng Li$^{4}$, 
    Xuanhua He$^{5}$, Jianhou Gan$^{2,*}$, Wenqi Ren$^{1,6,*}$ \\
    $^1$  Shenzhen Campus of Sun Yat-sen University 
    $^2$ Yunnan Normal University, Ministry of Education \\
    $^3$ Xidian University  
    $^4$ University of Science and Technology of China \\
    $^5$ The Hong Kong University of Science and Technology \\
    $^6$ MoE Key Laboratory of Information Technology
}

\begin{document}
\maketitle
\begingroup
\renewcommand{\thefootnote}{}
\footnotetext{$^\dagger$ Equal contribution. $^*$ Corresponding authors.}
\addtocounter{footnote}{-1}
\endgroup
\input{sec/0_abstract} 
\input{sec/1_intro}

\input{sec/2_formatting}
\input{sec/3_finalcopy}
\nocite{lai2016image,fan2024deep}
\balance % balance the final page (typically the references)
{\small
    \bibliographystyle{ieeenat_fullname}
    \bibliography{main}
}


\end{document}

%% file: preamble.tex
\usepackage{microtype}

%% file: sec/0_abstract.tex
\begin{abstract}

In this work,we propose a Multigrain-aware Semantic Prototype Scanning paradigm for pan-sharpening, built upon a high-order RWKV architecture and a tri-token prompting mechanism derived from
    semantic clustering. Specifically, our method contains three key components:
  1)Multigrain-aware Semantic Prototype Scanning. Although RWKV offers a efficient linear-complexity alternative to Transformers, its conventional bidirectional raster scanning is still semantic-agnostic and prone to positional bias. To address this issue, we introduce a semantic-driven scanning strategy that leverages locality-sensitive hashing to group
  semantically related regions and construct multi-grain semantic prototypes, enabling context-aware token reordering and more coherent global interaction.
  2)Tri-token Prompt Learning. We design a tri-token prompting mechanism consisting of a global token, cluster-derived prototype tokens, and a learnable register token. The global and
  prototype tokens provide complementary semantic priors for RWKV modeling, while the register token helps suppress noisy and artifact-prone intermediate representations. 
   3) Invertible Q-Shift. To counteract spatial details, we apply center difference convolution on the value pathway to inject high-frequency information, and introduce an invertible multi-scale Q-shift operation for efficient and lossless feature transformation without parameter-heavy receptive field expansion. 
  Experimental results demonstrate the superiority of our
  method.
\end{abstract}

%% file: sec/1_intro.tex
\section{Introduction}

\label{sec:intro}

\begin{figure*}[t]
	\centering
	\includegraphics[width=\textwidth]{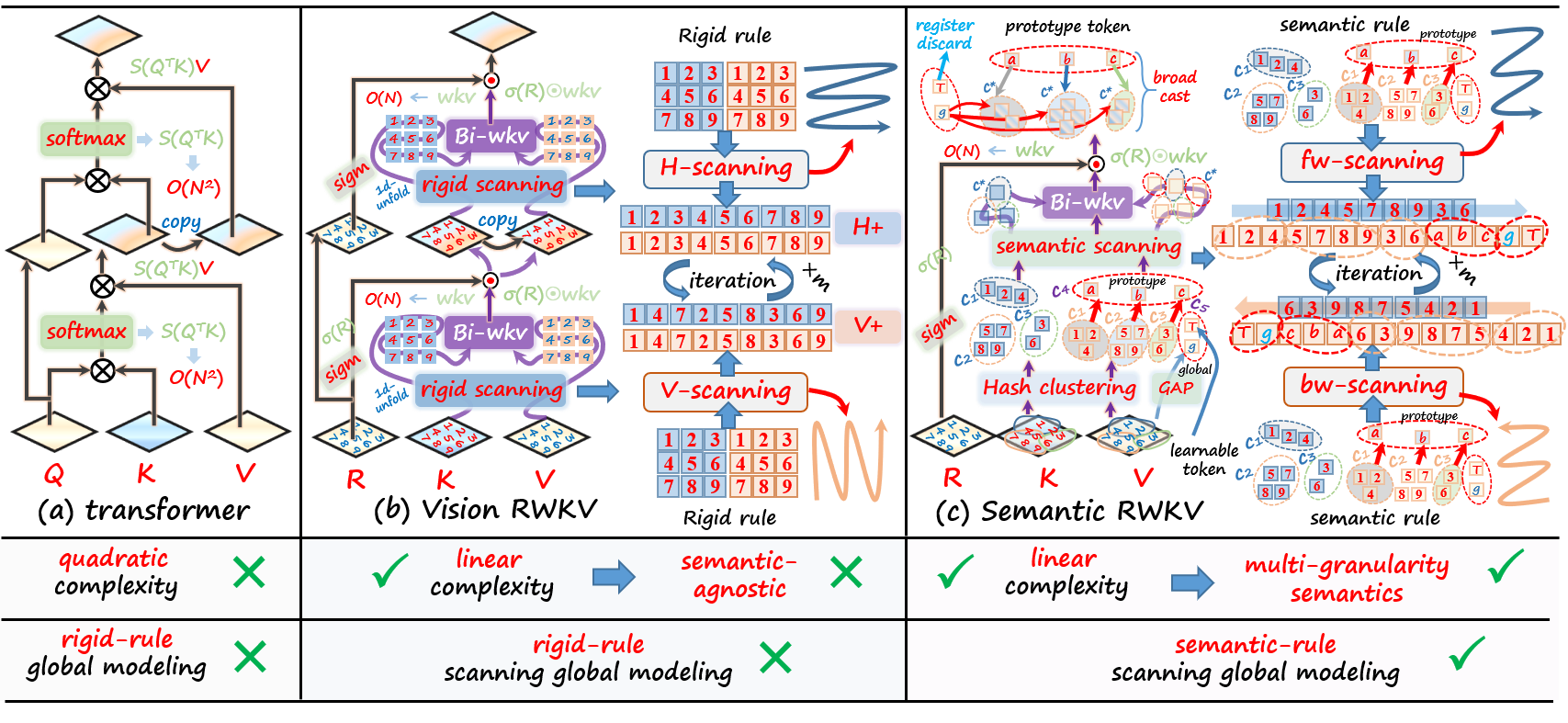}
	\caption{Comparison between the conventional Transformer, Vision RWKV, and our proposed Multigrain-aware Semantic Prototype Scanning paradigm.The typical (a) Transformer architecture suffers from quadratic complexity, which severely limits its applicability to high-resolution pan-sharpening tasks. Furthermore, (b) Vision RWKV, despite its linear efficiency, relies on a rigid bidirectional scanning strategy over $m$ iterations, leading to positional bias and an inability to adapt to semantic content. In contrast, (c) our proposed Multigrain-aware Semantic Prototype Scanning paradigm introduces a semantic-driven scanning path that dynamically prioritizes contextually coherent regions. Built upon a KV-sharing RWKV backbone for efficient global modeling, it is further enhanced by a tri-token prompting mechanism—derived from semantic clustering—to precisely guide the feature fusion process. This integrated approach achieves a more effective global receptive field, simultaneously eliminating memory constraints and semantic-agnostic modeling while ensuring computationally efficient and comprehensive spatial domain understanding.}
	\label{fig:teaser1}
\end{figure*}

\textbf{RWKV modeling.} Recent advances in pan-sharpening have witnessed the success of Transformer-based architectures in capturing long-range dependencies. However, their quadratic computational complexity among Key, Query and Value interaction $\mathbf{O} = \text{softmax}(\mathbf{Q}\mathbf{K}^T \rightarrow \mathcal{O}(N^2))\mathbf{V}$  remains a fundamental bottleneck for high-resolution remote sensing image. Characterized by linear complexity and inherent global modeling capability, the Vision RWKV architecture presents a promising alternative via its recurrent design, integrating  spatial mixer and  channel mixer
\begin{align}
  \textbf{\texttt{Spatial-mixer:}} wkv &= \textbf{\texttt{Bi-WKV}}(\mathbf{K}_s, \mathbf{V}_s), \\ \nonumber
  \mathbf{O}_s & = \texttt{Mapping}(\sigma(\mathbf{R}_s)\odot wkv), \\ \nonumber
  \texttt{\textbf{Channel-mixer:}} \mathbf{X}_c &= \textbf{\texttt{Q-Shift}}(\texttt{LN}(\mathbf{O}_s)), \\ \nonumber
 \mathbf{R}_c, \mathbf{V}_c & =\texttt{Mapping}, \gamma(\texttt{Mapping})(\mathbf{X}_c), \\  \nonumber
  \mathbf{O}_c &= \texttt{Mapping}((\sigma(\mathbf{R}_c) \odot \mathbf{V}_c)).  \nonumber
\end{align}
 Nonetheless, its attention mechanism, which employs a rigid-rule bidirectional scanning strategy $\textbf{\texttt{Bi-WKV}}(.)$, suffers from  positional bias and lacks semantic guidance, underscoring the need for further refinement to fully leverage its potential in pan-sharpening in \cref{fig:teaser1}. 

\textbf{Multigrain-aware Semantic Prototype Scanning.} Previous sequential scanning mechanism in RWKV follow a rigid raster pattern (\emph{i.e.}, semantic-agnostic horizontal and vertical sequencing) that fails to account for semantic coherence in image content. To address this limitation, we propose a semantic-driven scanning strategy that dynamically reorganizes processing order based on semantic relationships. By leveraging local hashing clustering, we generate semantic prototypes that guide the scanning path, ensuring semantically related regions are processed contiguously for more effective context modeling
\begin{align}
\textbf{\texttt{Local hashing:}}
& \mathbf{V}_{s}^{\texttt{index}}, \texttt{index} \leftarrow \textbf{\texttt{LSH}}(\mathbf{V}_s), \\
& \mathbf{K}_{s}^{\texttt{index}} \leftarrow  \textbf{\texttt{Cluster}}(\mathbf{K}_{s}, \texttt{index}), \nonumber 
\end{align}
Rooted in the semantic groups, we construct a set of informative tokens comprising semantic prototypes, a spatially averaged global token, and a learnable register token
\begin{align}
& \textbf{\texttt{Prototypes:}} \; \mathbf{P}_c = \frac{1}{|\mathcal{G}_c|} \sum_{i \in \mathcal{G}_c} \mathbf{V}_s^{(i)}, \; c = 1,\dots,C,  \nonumber \\
& \textbf{\texttt{Global:}} \;  \mathbf{g} = \frac{1}{T} \sum_{i=1}^T \mathbf{V}_s^{(i)},  
\textbf{\texttt{Register:}} \;  \mathbf{r} = \mathbf{W}_r \mathbf{V}_s^{\mathcal{I}} + \mathbf{b}_r, \nonumber
\end{align}
All the tokens appended to the unfolded sequence derived from $\mathbf{V}_{s}^{\texttt{index}}$ to enhance multigrain semantic awareness
\begin{align}
 \textbf{\texttt{Spatial-mixer:}} & \mathbf{V}_s  \leftarrow (\mathbf{V}_s, \mathbf{V}_{s}^{\mathtt{cls}}, \mathbf{V}_{s}^{\mathtt{g}}, \mathbf{V}_{s}^{\mathtt{r}}), \nonumber \\
& wkv = \textbf{\texttt{Bi-WKV}}(\mathbf{K}_s, \mathbf{V}_s), \nonumber \\ 
 & \mathbf{O}_s   = \texttt{Mapping}(\sigma(\mathbf{R}_s)\odot wkv), \nonumber \\
  & \mathbf{O}_{index}, \mathbf{O}_{cls}, \mathbf{O}_{g}, \mathbf{O}_{r}  \leftarrow \texttt{split}(\mathbf{O}_s). \nonumber
\end{align}
Upon obtaining the hierarchical-wise outputs, feature integration is performed through a structured broadcasting mechanism. This is achieved by propagating the global feature  $\mathbf{O}_{g}$ to all tokens in $\mathbf{O}_{index}$ , while simultaneously distributing each prototype-wise feature $\mathbf{O}_{cls}$ to its corresponding tokens in $\mathbf{O}_{index}$ where \text{index} equals \text{cls}. Concurrently, a register mechanism actively discards the captured noisy modal information $\texttt{discard}  \leftarrow  \mathbf{O}_{r}$ to purify the feature representation
\begin{align}
  \mathbf{O}_{index}  \frac{\text{broadcast}}{\leftarrow} \mathbf{O}_{g}, \mathbf{O}_{index}  \frac{\text{\texttt{index} = \texttt{cls}}}{\leftarrow} \mathbf{O}_{cls}. \nonumber
\end{align}

\textbf{High-order \& WKV-sharing/moment.} Despite the success of existing architectures, both second-order Transformers and RWKV models typically achieve higher-order interactions through the computationally expensive approach of stacking multiple blocks. This paradigm inherently requires recalculating the WKV (Weight-Key-Value) terms at each layer, resulting in considerable computational overhead. To address this limitation, we revisit the fundamental computational structure of WKV, which operates as a first-order function 
      $ \mathbf{O}_s  = \operatorname{sigmoid}(\mathbf{R}_s)\odot wkv,
     0< \operatorname{sigmoid}(\mathbf{R}_s(i)) < 1, \enspace \sum_{i}\operatorname{sigmoid}(\mathbf{R}_s(i))=1, \enspace \forall i \nonumber$
Mathematically, for any function $p(x)$ satisfying two constraints of $0 \le p(x) \le 1, \quad \sum_{x} p(x) = 1$ and acting as first-order statistic calculating, it equals to
\begin{equation}
      p(\mathbf{R}_{s}) \propto \operatorname{sigmoid}(\mathbf{R}_s(i)), 
\end{equation}
\begin{equation}
      \mathbf{O}_{s} = \int_{0}^{1}p(\mathbf{R}_{s})\mathbf{V}_{s}d\mathbf{v} \approx {\rm \textbf{E}}(\mathbf{v}_s),
\end{equation}
where $\rm \textbf{E}(\cdot)$ signifies first-order expectation calculating. Based on the insight in \cref{fig:core1},  we devise a lightweight higher-order RWKV architecture that moves beyond pure stacking, introducing a novel  WKV-sharing/moment strategy within/across groups 
\begin{align}
  \textbf{\texttt{WKV-sharing:}} 
  wkv_{(j)}^{(i-1)} &= \textbf{\texttt{WKV}}_{(j)}(\mathbf{K}_s, \mathbf{V}_s), \nonumber \\
  wkv_{(j)}^{(i)} & \leftarrow wkv_{(j)}^{(i-1)}, \nonumber \\
  \textbf{\texttt{WKV-moment:}} 
  wkv_{(j+1)}^{\frac{1}{2}} &= \textbf{\texttt{WKV}}_{(j+1)}(\mathbf{K}_s, \mathbf{V}_s), \nonumber \\
  wkv_{(j+1)}^{(1)} & \leftarrow \alpha \times wkv_{(j)}^{1} + wkv_{(j+1)}^{\frac{1}{2}}. \nonumber
\end{align}
where superscript $i$ and subscript $j$ denote the indices within and across RWKV groups, respectively.

\begin{figure*}[t]
	\centering
	\includegraphics[width=\textwidth]{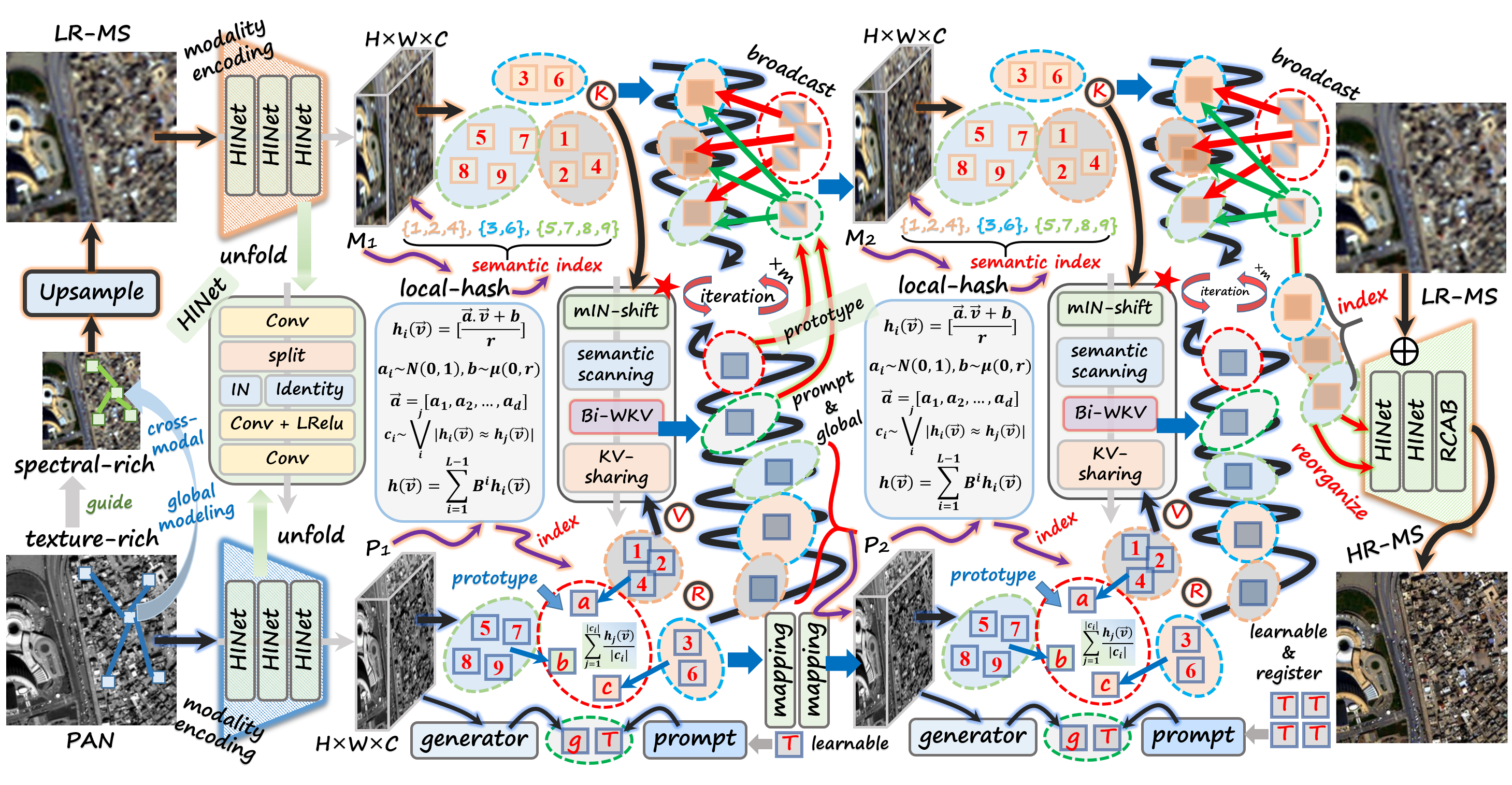}
	\caption{The Multigrain-aware Semantic Prototype Scanning architecture. It replaces standard recurrent scanning with a semantic-driven strategy, where the processing order is dynamically determined by clustering-derived prototypes. A novel tri-token prompt (global, prototype, register) guides the RWKV fusion process to be context-aware and artifact-robust.}
\label{fig:frameworkrs}
\end{figure*}

\textbf{INN-enhanced Q-shift.} Most current methods rely on parameter-intensive operations for spatial detail enhancement, leading to computational inefficiency. In contrast, our approach incorporates a multi-scale Q-shift operation within an invertible neural network. Notably, since Q-shift is functionally equivalent to a depth-wise $3 \times 3$ convolution $\textbf{\texttt{DWC}}_{3\times 3}  \leftarrow \text{Q-shift}$, it enables lossless feature transformation with high computational efficiency. This design preserves high-frequency details through an elegant mathematical formulation, avoiding excessive parameter growth.
Furthermore, recognizing that RWKV belongs to the linear attention family—interpreted as performing low-pass filtering—we incorporate  central difference convolution ($\textbf{\texttt{CDC}}$) to enhance the model's ability to capture high-frequency information
$\textbf{\texttt{CDC:}}  \mathbf{O}_h =  \mathbf{O}_s + \textbf{\texttt{CDC}}(\mathbf{K}_s)$.

\textbf{Solutions.} In this work, we propose a Multigrain-aware Semantic Prototype Scanning paradigm for pan-sharpening, built upon a KV-sharing RWKV architecture for efficient global modeling. Our framework integrates a semantic-driven scanning strategy to eliminate positional biases inherent in traditional RWKV, along with a tri-token prompting mechanism derived from semantic clustering to guide the fusion process. The tri-token mechanism comprises: (i) a global token capturing holistic context, (ii) cluster-derived prototype tokens representing distinct semantic regions, and (iii) a learnable token register that dynamically suppresses feature artifacts. Additionally, we introduce a dual-path enhancement design incorporating center difference convolution and an invertible neural network with multi-scale Q-shift operation, enabling high-frequency detail preservation through efficient, lossless feature transformation. Extensive experiments demonstrate that our method achieves superior performance in both spatial-spectral detail reconstruction.

The contributions of this work are as follows.
\begin{itemize} 
\item Semantic-guided RWKV: We propose a novel multigrain-aware semantic prototype scanning strategy that replaces conventional fixed-pattern RWKV scanning with semantic-driven processing. By leveraging local hashing-based clustering to organize scanning order according to semantic coherence, our approach enables context-aware global modeling while maintaining linear complexity.

\item Tri-token Prompting Mechanism: We introduce a novel tri-token prompting framework comprising globally-sourced tokens, cluster-derived prototype tokens, and learnable register tokens. This design dynamically guides the fusion process through semantic-aware conditioning while effectively suppressing feature artifacts through adaptive regularization.

\item High-frequency Enhancement: We develop  dual-path enhancement strategy combining center difference convolution with  invertible neural network featuring multi-scale Q-shift. This design enables lossless feature transformation and high-frequency detail preservation without parameter-intensive receptive field expansion.
  
\end{itemize}

\section{Related work}

Existing pan-sharpening approaches can be broadly categorized into component substitution (CS), multi-resolution analysis (MRA), and variational optimization (VO) frameworks. CS methods replace spatial components of multispectral (MS) images with panchromatic (PAN) details, exemplified by intensity-hue-saturation (IHS), principal component analysis (PCA), Brovey fusion, and Gram–Schmidt (GS) spectral sharpening \cite{carper1990use, gillespie1987color}. While computationally efficient, they often introduce spectral distortion. Refinements like nonlinear IHS (NIHS) and adaptive GS (GSA) better preserve spectral traits \cite{NIHS2016, GSA2007}, yet still struggle with spatial fidelity in complex terrains.
VO-based methods formulate pan-sharpening as a constrained optimization problem to preserve spectral properties. Modern VO techniques employ advanced regularizers such as gradient sparsity (SIRF), and group-based low-rank priors (ADMM) \cite{SIRF2015, LGC2019, ADMM2021}. Although theoretically sound, VO methods require careful parameter tuning and may not fully capture complex structures, limiting scalability for large-scale remote sensing applications\cite{wang2024omnidirectional}.
The advent of CNNs has significantly advanced pan-sharpening through hierarchical feature learning. Pioneering work by Masi et al. \cite{pnn} demonstrated CNNs' superiority in learning mappings between low- and high-resolution images. Subsequent innovations include residual connections for deeper architectures \cite{pannet, resnet}, multi-scale modules for contextual capture \cite{msdcnn,rap-sr}, and enriched input representations combining local and global cues \cite{srppnn}. Recent trends incorporate domain knowledge via physics-aware designs, such as modality-specific priors \cite{gppnn}, unrolled optimization algorithms \cite{9165231,qin2024enhanced}, and hybrid models merging variational methods with deep networks \cite{ma3, dengVO,quan2024siamese}. 
Despite these advances, potential of linear-complexity  RWKV remains unexplored in pan-sharpening, suggesting a promising direction for future research.

%% file: sec/2_formatting.tex
\section{Method}
 The proposed RWKV framework in \cref{fig:frameworkrs} takes as input  PAN image $\mathbf{I}_{\mathcal{P}} \in \mathbb{R}^{\rm H \times W \times 1}$ and a low-resolution multispectral image $\mathbf{I}_{\mathcal{M}} \in \mathbb{R}^{\rm h \times w \times C}$. These inputs are processed through dual modality-specific encoders to extract representative features
\begin{equation}
\mathbf{F}_{\mathcal{P}} = E_{\mathcal{P}}(\mathbf{I}_{\mathcal{P}}), \quad \mathbf{F}_{\mathcal{M}} = E_{\mathcal{M}}(\text{Up}(\mathbf{I}_{\mathcal{M}})),
\end{equation}
where $E_{\mathcal{P}}(\cdot)$ and $E_{\mathcal{M}}(\cdot)$ denote the encoding functions for PAN and multispectral modalities, respectively. Each encoder incorporates half-instance normalization as its fundamental component. The resulting modality-aware features are then progressively transformed through $L$ layers of the Multigrain-aware Semantic Prototype Scanning and Tri-token Prompt Learning embraced high-order RWKV module  with the initial features $\mathbf{F}^{(0)}_{\mathcal{P}} = \mathbf{F}_{\mathcal{P}}, \mathbf{F}^{(0)}_{\mathcal{M}} = \mathbf{F}_{\mathcal{M}}$
\begin{equation}
\mathbf{F}^{(i)}_{\mathcal{P}}, \mathbf{F}^{(i)}_{\mathcal{M}} = \text{MTRWKV}^{(i)}\left(\mathbf{F}^{(i-1)}_{\mathcal{P}}, \mathbf{F}^{(i-1)}_{\mathcal{M}}\right), i = 1, \dots, L
\end{equation}
where $\text{MTRWKV}^{(i)}$ represents the $i$-th high-order interaction block. The final fused image $\mathbf{I}_{\mathcal{F}} \in \mathbb{R}^{\rm H \times W \times C}$ is reconstructed through a decoder $D_{\mathcal{C}}(\cdot)$ 
\begin{equation}
\mathbf{I}_{\mathcal{F}} = D_{\mathcal{C}}(\mathbf{F}^{(L)}_{\mathcal{H}}) + \text{Up}(\mathbf{I}_{\mathcal{M}}),
\end{equation}
where $\mathbf{F}^{(L)}_{\mathcal{H}}$ denotes aggregated high-order features from final MTRWKV, and $\text{Up}(\cdot)$ represents  upsampling operation.

\begin{figure*}[t]
	\centering
	\includegraphics[width=\textwidth]{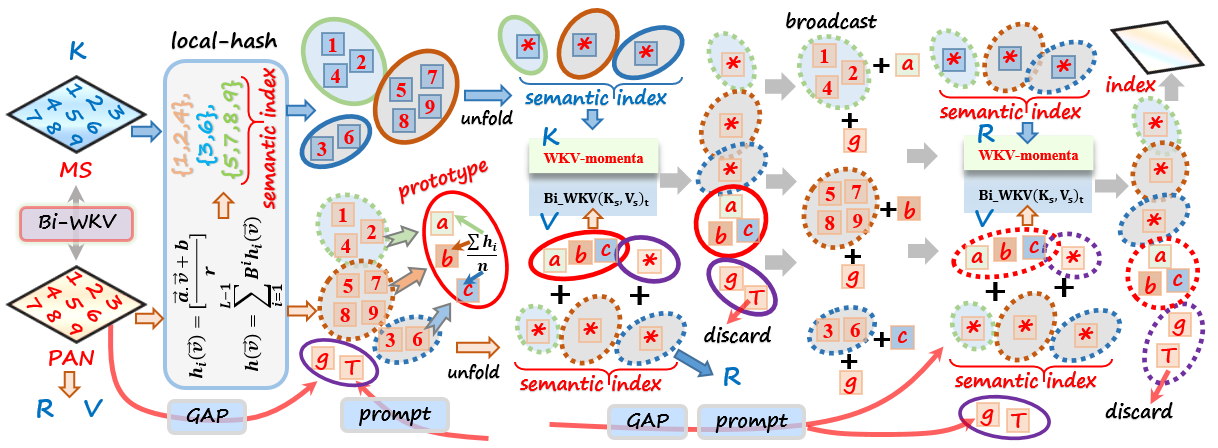}
	\caption{Semantic-guided scanning MTRWKV. It replaces the standard raster scan with a semantic order derived from feature clustering, processes the sequence with a prompted Bi-WKV, and then restores original spatial arrangement through an inverse reorganization.}
\label{fig:core1}
\end{figure*}

\subsection{Multigrain-aware semantic scanning and tri-token prompt} \label{main}
\textbf{Preliminaries of RWKV.} The Vision RWKV architecture \cite{duan2025RWKV} comprises two core components: a spatial mixer for global attention and a channel mixer for feature fusion. Given input features $\mathbf{F}^{(i)}_{\mathcal{M}}$ and $\mathbf{F}^{(i)}_{\mathcal{P}}$ flattened into sequences $\mathbf{X}_\mathcal{M}, \mathbf{X}_\mathcal{P} \in \mathbb{R}^{\rm T \times C}$ where $\rm T = H \times W$, the spatial mixer first applies layer normalization and token shifting
$\mathbf{R}_s = \mathbf{X}_\mathcal{M} \mathbf{W}_R, 
  \mathbf{K}_s = \mathbf{X}_\mathcal{P} \mathbf{W}_K, 
 \mathbf{V}_s = \mathbf{X}_\mathcal{P} \mathbf{W}_V, 
 wkv = \texttt{Bi-WKV}(\mathbf{K}_s, \mathbf{V}_s),
 \mathbf{O}_s = \mathbf{W}_O(\sigma(\mathbf{R}_s) \odot wkv)$,
where $\mathbf{W}_R, \mathbf{W}_K, \mathbf{W}_V$ are linear projection weights. Bi-WKV calculation employs the bidirectional scanning mechanism with $m$ iterations
\begin{align}
wkv_t &= \frac{\sum_{i \neq t} e^{-(|t-i|-1)w/T + k_i}v_i + e^{u + k_t}v_t}
{\sum_{i \neq t} e^{-(|t-i|-1)w/T + k_i} + e^{u + k_t}}, \nonumber \\
wkv &= \text{Re-WKV}_{(m)}(\mathbf{K}_s, \mathbf{V}_s),
\end{align}
where $u,w$ are hyperparameters. The spatial mixer's output $\mathbf{O}_{sf}$ is then passed to the channel mixer along with $F^{(i)}_{\mathcal{P}}$
\begin{align}
\mathbf{O}_{sc} &= \texttt{Concat}(\mathbf{O}_{sf}, \mathbf{F}^{(i)}_{\mathcal{P}}), \nonumber \\
\mathbf{X}_c &= \texttt{Omni-Shift}(\texttt{LN}(\mathbf{O}_{sc})), \nonumber \\
\mathbf{R}_c, \mathbf{V}_c &= \mathbf{W}_1(\mathbf{O}_{sc}), \gamma (\mathbf{W}_2(\mathbf{O}_{sc})), \nonumber \\
\mathbf{O}_c &= \mathbf{W}_3(\sigma(\mathbf{R}_c) \odot \mathbf{V}_c),
\end{align}
where $\gamma$ is a scaling factor and $\mathbf{W}_1, \mathbf{W}_2, \mathbf{W}_3$ are MLP projections. 
Nonetheless, its attention mechanism, which employs a rigid-rule bidirectional scanning strategy $\textbf{\texttt{Bi-WKV}}(.)$, suffers from  positional bias and lacks semantic guidance, underscoring the need for further refinement to fully leverage its potential in pan-sharpening in \cref{fig:teaser1}. 

\textbf{Local Sensitive Hashing.}
We employ Locality-Sensitive Hashing  to adaptively cluster queries based on their Euclidean proximity. LSH ensures that nearby vectors receive identical hash values with high probability, while distant vectors are hashed differently. By tuning hash function parameters and the number of hashing rounds, vectors within a distance threshold $\varepsilon$ are grouped into the same hash bucket with probability exceeding $p$. We adopt Exact Euclidean LSH as
$h(\vec{v}) = \left\lfloor \frac{\vec{a} \cdot \vec{v} + b}{r} \right\rfloor$,
where $h: \mathbb{R}^d \rightarrow \mathbb{Z}$, $r$ is a hyper-parameter controlling bucket width, $\vec{a} = (a_1, \dots, a_d)$ with $a_i \sim \mathcal{N}(0,1)$, and $b \sim \mathcal{U}(0,r)$. To enhance robustness, we apply $L$ independent hashing rounds. The final hash value combines these rounds via
$h(\vec{v}) = \sum_{i=0}^{L-1} B^i h_i(\vec{v})$,
where each $h_i$ uses independently sampled $\vec{a}_i$ and $b_i$. Each $h_i$ corresponds to a set of parallel hyperplanes with normal vector $\vec{a}_i$ and offset $b_i$. The spacing between hyperplanes is regulated by $r$ and $L$ hash functions partition the space into cells, with vectors in the same cell sharing a hash value. 

The cluster assignment $G_i$ for query $Q_i$ is determined by its hash value. The prototype for the $j$-th cluster is computed as
\begin{equation}
P_j = \frac{\sum_{i: G_i = j} Q_i}{\sum_{i: G_i = j} 1}.
\end{equation}
The prototype set $P \in \mathbb{R}^{C \times D_k}$ compactly represents the query distribution, where $C$ is the number of clusters. To refine the prototype representation, we introduce weighting scheme based on query density within each cluster
\begin{equation}
w_j = \frac{\log\left(1 + \sum_{i: G_i = j} 1\right)}{\sum_{k=1}^C \log\left(1 + \sum_{i: G_i = k} 1\right)},
\end{equation}
yielding the weighted prototype $\tilde{P}_j = w_j \cdot P_j$ and emphasizing clusters with higher query density.

\textbf{Multigrain-aware Semantic Prototype Scanning.} 
Traditional sequential scanning mechanisms in RWKV follow rigid raster patterns  that fail to account for semantic coherence in image content. To overcome this limitation, we introduce a semantic-driven scanning strategy that dynamically reorganizes the processing sequence based on semantic relationships in \cref{fig:core1}. By leveraging locality-sensitive hashing clustering, we generate semantic prototypes that guide the scanning path, ensuring semantically related regions are processed contiguously for enhanced contextual modeling. \textbf{(1) Semantic-guided Token Reorganization.} The semantic reorganization begins with locality-sensitive hashing applied to the value matrix
\begin{align}
\textbf{\texttt{LSH:}} \; & \mathcal{I}, \{\mathcal{G}_c\}_{c=1}^C = \texttt{LSH}(\mathbf{V}_s),  \\
& \mathbf{V}_s^{\mathcal{I}} = \texttt{Reorder}(\mathbf{V}_s, \mathcal{I}), 
\mathbf{K}_s^{\mathcal{I}} = \texttt{Reorder}(\mathbf{K}_s, \mathcal{I}),  \nonumber
\end{align}
where $\mathcal{I}$ denotes the semantic-aware indexing, $\{\mathcal{G}_c\}$ represents semantic groups, and $C$ is the cluster count.
\textbf{(2) Multigrain Token Construction.} We construct a comprehensive token set comprising semantic prototypes, global context, and adaptive registers
\begin{align}
& \textbf{\texttt{Prototypes:}} \; \mathbf{P}_c = \frac{1}{|\mathcal{G}_c|} \sum_{i \in \mathcal{G}_c} \mathbf{V}_s^{(i)}, \; c = 1,\dots,C,  \nonumber \\
& \textbf{\texttt{Global:}} \;  \mathbf{g} = \frac{1}{T} \sum_{i=1}^T \mathbf{V}_s^{(i)},  
\textbf{\texttt{Register:}} \;  \mathbf{r} = \mathbf{W}_r \mathbf{V}_s^{\mathcal{I}} + \mathbf{b}_r, \nonumber
\end{align}
where $\mathbf{W}_r$ and $\mathbf{b}_r$ are learnable parameters.
\textbf{(3) Semantic-aware Processing.} The enhanced token sequence undergoes semantic-aware transformation
\begin{align}
\mathbf{V}_s^{\text{enh}} &= [\mathbf{V}_s^{\mathcal{I}}; \mathbf{P}_1, \dots, \mathbf{P}_C; \mathbf{g}; \mathbf{r}], \\
wkv &= \texttt{Bi-WKV}(\mathbf{K}_s, \mathbf{V}_s^{\text{enh}}), \\
\mathbf{O}_s &= \mathbf{W}_o (\sigma(\mathbf{R}_s) \odot wkv),
\end{align}
where $\mathbf{W}_o$ denotes the output projection.
\textbf{(4) Hierarchical Feature Integration} The processed outputs are decomposed and integrated through structured broadcasting:
\begin{align}
& \mathbf{O}^{\mathcal{I}}, \{\mathbf{O}_c^{\texttt{proto}}\}_{c=1}^C, \mathbf{O}^{\texttt{global}}, \mathbf{O}^{\texttt{reg}} = \texttt{Split}(\mathbf{O}_s), \nonumber \\
& \mathbf{O}_c^{\mathcal{I}} = \mathbf{O}^{\mathcal{I}}[\mathcal{G}_c] + \mathbf{O}_c^{\texttt{proto}} + \mathbf{O}^{\texttt{global}}, \nonumber \\
& \mathbf{O}^{\texttt{final}} = \texttt{Concat}(\{\mathbf{O}_c^{\mathcal{I}}\}_{c=1}^C),
\end{align}
where $\mathbf{O}^{\texttt{reg}}$ is discarded to suppress noise. This multi-grain integration ensures comprehensive semantic awareness while maintaining computational efficiency.

\begin{figure}[t]
	\centering
	\includegraphics[width=0.9\columnwidth]{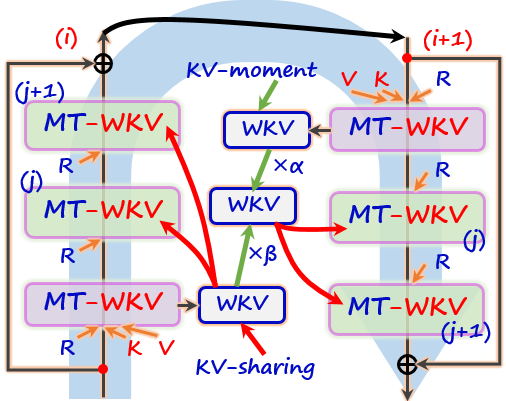}
	\caption{WKV-moment/sharing mechanism.}
	\label{fig:graphagg}
\end{figure}

\textbf{High-order \& WKV-sharing/moment.} 
Existing architectures, including second-order Transformers and RWKV models, typically achieve higher-order interactions through computationally expensive block stacking. This paradigm requires recalculating Weight-Key-Value (WKV) terms at each layer, leading to significant computational overhead.  \textbf{(1) Revisiting WKV as First-order Expectation.}
We analyze the fundamental computational structure of WKV, which operates as a first-order weighting function:
\begin{align}
\mathbf{O}_s &= \sigma(\mathbf{R}_s) \odot \text{WKV}(\mathbf{K}_s, \mathbf{V}_s), \\
0 &< \sigma(\mathbf{R}_s(i)) < 1, \quad \sum_{i} \sigma(\mathbf{R}_s(i)) = 1, \quad \forall i,
\end{align}
where $\sigma(\cdot)$ denotes the sigmoid function. Mathematically, for any probability mass function $p(x)$ satisfying
$0 \leq p(x) \leq 1, \quad \sum_{x} p(x) = 1$,
we can express WKV as
\begin{align}
p(\mathbf{R}_s) \propto \sigma(\mathbf{R}_s), 
\mathbf{O}_s = \mathbb{E}_{\mathbf{v} \sim p(\mathbf{R}_s)}[\mathbf{v}] \approx \mathbb{E}[\mathbf{V}_s],
\end{align}
where $\mathbb{E}[\cdot]$ represents the first-order expectation operation.
\textbf{(2) WKV-Sharing and Moment Strategy.}
Based on this insight (\cref{fig:graphagg}), we propose a lightweight higher-order RWKV architecture that moves beyond pure stacking. Our approach introduces two complementary strategies: (a) WKV-Sharing Across Layers.
We share WKV computations across consecutive layers to reduce redundancy:
\begin{align}
\texttt{WKV-Sharing:} \quad & wkv_{(j)}^{(i-1)} = \text{WKV}_{(j)}(\mathbf{K}_s, \mathbf{V}_s), \\
& wkv_{(j)}^{(i)} \leftarrow wkv_{(j)}^{(i-1)},
\end{align}
where superscript $i$ denotes the layer index within a group, and subscript $j$ represents the group index. (b) Moment-based Feature Propagation.
We employ a moment-based mechanism for cross-group feature enhancement:
\begin{align}
& \texttt{WKV-Moment:}   wkv_{(j+1)}^{\frac{1}{2}} = \text{WKV}_{(j+1)}(\mathbf{K}_s, \mathbf{V}_s), \\
& wkv_{(j+1)}^{(1)} \leftarrow \alpha \cdot wkv_{(j)}^{1} + (1-\alpha) \cdot wkv_{(j+1)}^{\frac{1}{2}},
\end{align}
where $\alpha \in [0,1]$ is a learnable momentum parameter.

\begin{figure}[t]
	\centering
	\includegraphics[width=\columnwidth]{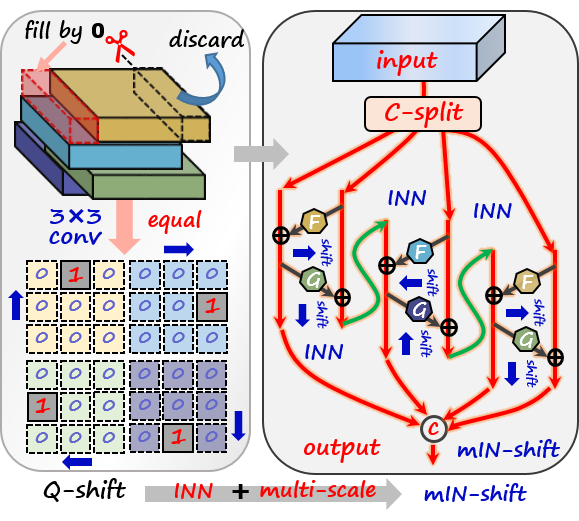}
	\caption{Invertible neural network-based Q-shift.}
	\label{fig:shift}
\end{figure}

\begin{table*}[t]
	\centering
	% \normalsize
% 	\renewcommand\arraystretch{1.2}
	\renewcommand{\tabcolsep}{1pt} % adjust horizontal space
\renewcommand{\arraystretch}{1}
\caption{Comparison on the WordView-II, WordView-III and GaoFen2 datasets. }
	\resizebox{0.966\textwidth}{!}{
	\begin{tabular}{c|cccc|cccc|cccc}
		\hline
  & \multicolumn{4}{c|}{WordView-II}                                                      
& \multicolumn{4}{c}{WordView-III}  & \multicolumn{4}{c}{GaoFen2}                                                             \\ \cline{2-13} 
		\multirow{-2}{*}{Method} & PSNR $\uparrow$& SSIM $\uparrow$ &SAM $\downarrow$ & ERGAS $\downarrow$   & PSNR $\uparrow$& SSIM $\uparrow$ &SAM $\downarrow$ & ERGAS $\downarrow$   & PSNR $\uparrow$& SSIM $\uparrow$ &SAM $\downarrow$ & ERGAS $\downarrow$ \\
		\hline
		SFIM   &34.1297& 0.8975& 0.0439& 2.3449 &21.8212& 0.5457&0.1208& 8.9730&36.9060& 0.8882& 0.0318& 1.7398 \\
		GS      & 35.6376& 0.9176& 0.0423& 1.8774 & 22.5608& 0.5470& 0.1217& 8.2433 & 37.2260 & 0.9034& 0.0309& 1.6736 \\
		
		Brovey  & 35.8646& 0.9216& 0.0403& 1.8238& 22.5060& 0.5466& 0.1159& 8.2331 & 37.7974& 0.9026& 0.0218& 1.3720 \\
		
		IHS   & 35.2962& 0.9027& 0.0461& 2.0278  & 22.5579& 0.5354& 0.1266& 8.3616& 38.1754& 0.9100& 0.0243& 1.5336 \\

		GFPCA    & 34.558& 0.9038& 0.0488& 2.1400 & 22.3400& 0.4826& 0.1294& 8.3964& 37.9443& 0.9204& 0.0314& 1.5604 \\ \hline
	
		PNN   &40.755 & 0.9624 & 0.0259 & 1.0646  &29.9418& 0.9121& 0.0824& 3.3206& 43.1208& 0.9704& 0.0172& 0.8528  \\
		
		PANNet   & 40.8176& 0.9626& 0.0257& 1.0557  & 29.6840& 0.9072& 0.0851& 3.4263&  43.0659& 0.9685& 0.0178& 0.8577 \\
		
		MSDCNN  &41.3355& 0.9664& 0.0242& 0.9940  &30.3038& 0.9184& 0.0782& 3.1884& 45.6874& 0.9827& 0.0135& 0.6389 \\
		
		SRPPNN  &41.4538& 0.9679& 0.0233& 0.9899 &30.4346& 0.9202& 0.0770& 3.1553& 47.1998& 0.9877& 0.0106& 0.5586  \\
		
		GPPNN &41.1622& 0.9684& 0.0244& 1.0315  &30.1785& 0.9175& 0.0776& 3.2593& 44.2145& 0.9815& 0.0137& 0.7361 \\ 
  
    MutNet &41.6773& 0.9705& 0.0224& 0.9519 &30.4907& 0.9223& 0.0749& 3.1125& 47.3042& 0.9892& 0.0102& 0.5481 \\ 
		
		SFINet& 41.7244  & 0.9725& 0.0220 & 0.9506  &30.5971 &0.9236&0.0741& 3.0798& 47.4712 & 0.9901 & 0.0102 & 0.5462  \\

  % SHIP &\textbf{41.8169}  &\textbf{0.9733}& \textbf{0.0220} &\textbf{0.9481}& \textbf{0.9737}& \textbf{0.7768}\\
  
  PanFlowNet & 41.8548  & 0.9712 & 0.0224 & 0.9335   & 30.4873 & 0.9221 & 0.0751 & 3.1142  & 47.2533 & 0.9884 & 0.0103 & 0.5512   \\ \hline

    Ours &\textbf{42.3751} & \textbf{0.9737} & \textbf{0.0208} & \textbf{0.8816}  & \textbf{31.3113} & \textbf{0.9319} & \textbf{0.0685} & \textbf{2.8361} &\textbf{47.8941} &\textbf{0.9901} &\textbf{0.0097} &\textbf{0.5115} \\
		
		\hline
	\end{tabular}}%
	\label{t1}%
\end{table*}%

\textbf{INN-enhanced Q-shift.} 
Most current pan-sharpening methods rely on parameter-intensive operations for spatial detail enhancement, leading to computational inefficiency and scalability limitations. \textbf{(1) Invertible Neural Network with Multi-scale Q-Shift.}
To address this limitation, we incorporate a multi-scale Q-shift operation within an invertible neural network (INN) framework in \cref{fig:shift}. The Q-shift operation is functionally equivalent to a depth-wise $3 \times 3$ convolution $\texttt{Q-shift} \equiv \texttt{DWC}_{3\times3}$, but achieves lossless feature transformation. Our INN architecture processes input features through a multi-stage reversible transformation:
\begin{align}
\mathbf{X}_1, \mathbf{X}_2, \mathbf{X}_3, \mathbf{X}_4 = \texttt{Split}(\mathbf{X}), \nonumber \\
\textbf{\texttt{INN:}} \mathbf{Y}_1 = \mathbf{X}_{1} + \mathcal{F}_1(\mathbf{X}_{2}), 
\mathbf{Z}_2 = \mathcal{G}_1(\mathbf{Y}_1)  + \mathbf{X}_{2}, \nonumber \\
\mathbf{Y}_2 = \mathbf{Z}_{2} + \mathcal{F}_2(\mathbf{X}_{3}), \mathbf{Z}_3 = \mathcal{G}_2(\mathbf{Y}_2)  + \mathbf{X}_{3}, \nonumber \\
\mathbf{Y}_3 = \mathbf{Z}_{3} + \mathcal{F}_3(\mathbf{X}_{4}), \mathbf{Z}_4 = \mathcal{G}_3(\mathbf{Y}_3)  + \mathbf{X}_{4}, \nonumber \\
\textbf{\texttt{Aggregation:}} \mathbf{Y}_o = \texttt{Concate}(\mathbf{Y}_1, \mathbf{Y}_2, \mathbf{Y}_3, \mathbf{Z}_4) \nonumber
\end{align}
where $\mathcal{F}_i$ and $\mathcal{G}_i$ represent spatial shift operators along different directions, implemented via Q-shift operations. \textbf{(2) High-Frequency Enhancement via Central Difference Convolution.}
Recognizing that RWKV's linear attention mechanism performs token-level filtering (effectively low-pass filtering), we incorporate central difference convolution (CDC) to enhance high-frequency information capture:
\begin{equation}
\text{CDC:} \quad \mathbf{O}_h = \mathbf{O}_s + \text{CDC}(\mathbf{K}_s).
\end{equation}
CDC emphasizes gradient information by computing differences between central and surrounding pixels
\begin{equation}
\text{CDC}(\mathbf{K}_s) = \sum_{i=-1}^{1}\sum_{j=-1}^{1} w_{ij} \cdot [\mathbf{K}_s(x,y) - \mathbf{K}_s(x+i,y+j)]
\end{equation}
where $w_{ij}$ are learnable parameters that emphasize high-frequency components.

\begin{table*}[htbp]
\centering
\caption{Performance comparison of different configurations on multiple datasets.}
\resizebox{\textwidth}{!}{
\begin{tabular}{cccccccccccc}
\toprule
\multicolumn{4}{c}{\textbf{Config}} & \multicolumn{4}{c}{\textbf{WV2}} & \multicolumn{4}{c}{\textbf{GF2}} \\
\cmidrule(lr){1-4} \cmidrule(lr){5-8} \cmidrule(lr){9-12} 
\textbf{Momentum} & \textbf{Avg Token} & \textbf{Learn Token} & \textbf{Prototype} & \textbf{PSNR} & \textbf{SSIM} & \textbf{SAM} & \textbf{ERGAS} & \textbf{PSNR} & \textbf{SSIM} & \textbf{SAM} & \textbf{ERGAS} \\
\midrule
$\checkmark$ & $\times$ & $\times$ & $\times$ & 42.0364 & 0.9716 & 0.0216 & 0.9177 & 47.5367 & 0.9893 & 0.0103 & 0.5319  \\
$\checkmark$ & $\checkmark$ & $\times$ & $\times$ & 42.1360 & 0.9723 & 0.0214 & 0.9031 & 47.6217 & 0.9896 & 0.0098 & 0.5259  \\
$\checkmark$ & $\checkmark$ & $\checkmark$ & $\times$ & 42.2192 & 0.9730 & 0.0210 & 0.8954 & 47.6681 & 0.9897 & 0.0099 & 0.5248  \\
$\times$ & $\checkmark$ & $\checkmark$ & $\checkmark$ & 42.2800 & 0.9733 & 0.0209 & 0.8912 & 47.7976 & 0.9899 & 0.0099 & 0.5186  \\
$\checkmark$ & $\checkmark$ & $\checkmark$ & $\checkmark$ & 42.3750 & 0.9737 & 0.0208 & 0.8816 & 47.8941 & 0.9901 & 0.0097 & 0.5115  \\
\bottomrule
\end{tabular}}
\label{abla}
\end{table*}

\begin{figure*}[t]
	\centering
	\includegraphics[width=0.94\textwidth]{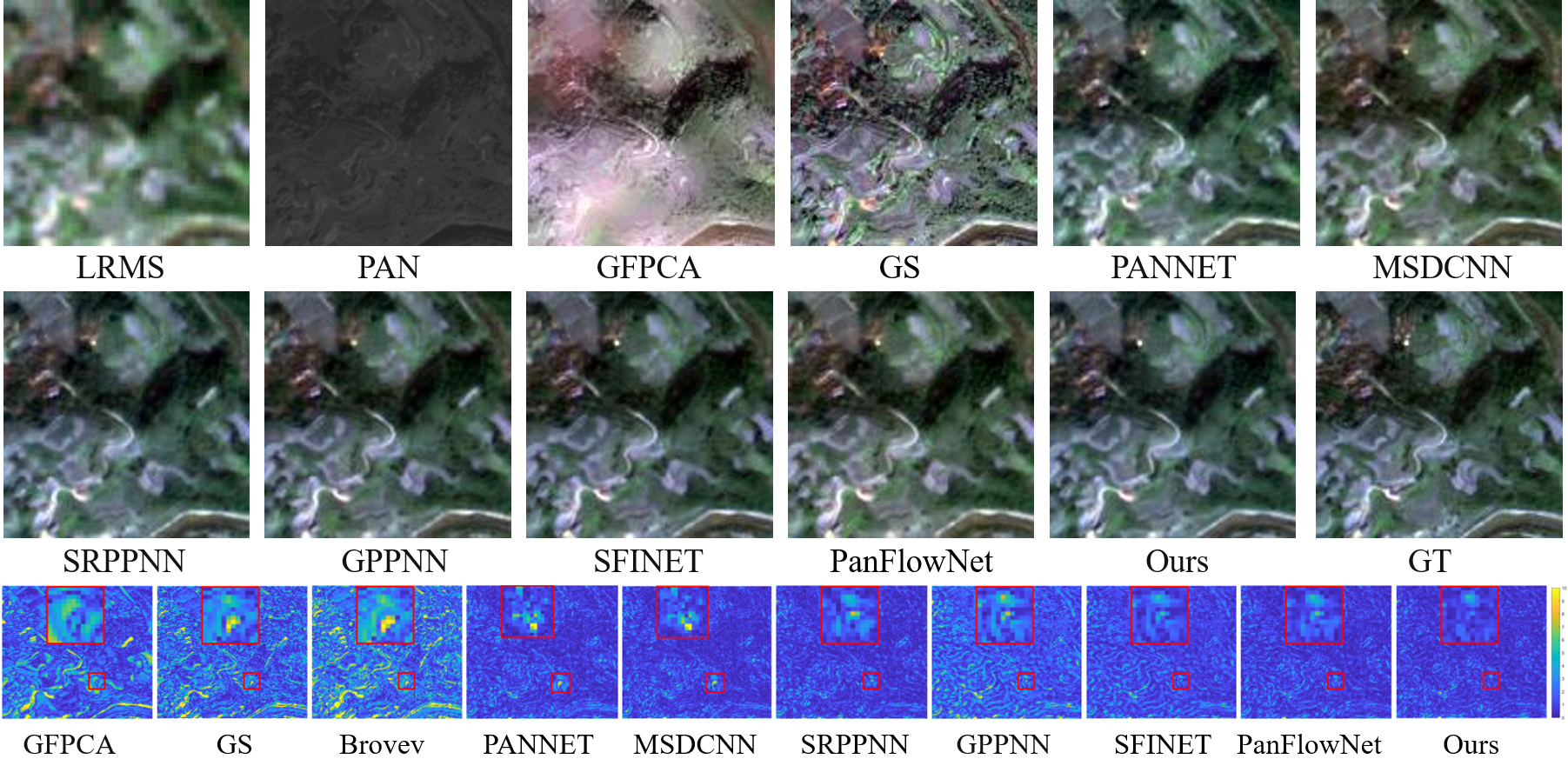}
	\caption{Visual comparison evidenced by the MSE residue map between ground truth and prediction.}
	\label{vis}
\end{figure*}

\section{Experiments over pan-sharpening}
To evaluate the performance, we conduct comparative analysis against pan-sharpening. The traditional methods included SFIM~\cite{SFIM}, Brovey~\cite{Brovey}, GS~\cite{GS}, IHS~\cite{IHS}, and GFPCA~\cite{GFPPCA}. Additionally, we include various deep learning-based techniques, such as PNN~\cite{pnn}, PANNET~\cite{pannet}, MSDCNN~\cite{msdcnn}, SRPPNN~\cite{srppnn}, GPPNN~\cite{gppnn}, MutNet~\cite{Zhou_2022_CVPR1},  SFINet~\cite{zhou2022spatial}, and PanFlowNet~\cite{Yang_2023_ICCV}. During the training phase, we implemented the Adam optimizer for a total of 500 epochs, utilizing a batch size of 4 and an initial learning rate set at 
$5 \times 10^{-4}$. The learning rate was halved every 100 epochs to facilitate stable convergence over satellite dataset. 

\textbf{Comparisons with SOTA.}
To assess the performance, we employed a diverse set of metrics, with the results systematically presented in \cref{t1}. These results highlight the outstanding performance of our techniques, clearly demonstrating their superiority over benchmark algorithms across all evaluation criteria. Specifically, our methods achieve PSNR improvements of 0.52 dB, 0.83 dB, and 0.64 dB relative to the highest-performing algorithm for the WorldView-II, GaoFen2, and WorldView-III datasets, respectively. Visual comparison in \cref{vis} also supports the above claim.

\textbf{Effect of Token Processing Strategies:} We systematically evaluate the impact of four key components by progressively enabling them across configurations (I)–(V) in \cref{abla}. The analysis reveals distinct synergistic effects among these components.
The combination of Average Token and Learn Token mechanisms (Configuration III) demonstrates significant improvements over individual implementations. When used alone, Average Token (Configuration II) provides moderate gains, but the synergistic integration with Learn Token (Configuration III) yields more substantial performance enhancements across all metrics.

\textbf{Effect of Momentum Mechanisms:}
The transition from Configuration III to V highlights the critical role of Momentum and Prototype mechanisms. Configuration IV shows that adding Prototype learning without Momentum provides limited benefits. However, the complete system (Configuration V) with all four components activated – Momentum, Average Token, Learn Token, and Prototype – achieves the optimal performance (PSNR: 42.3750, SSIM: 0.9737 on WV2; PSNR: 47.8941, SSIM: 0.9901 on GF2), confirming the importance of integrated framework.

% \textbf{Effect of Prompt-learning Tri-token:}
% We systematically evaluate semantic scanning and tri-token prompting by progressively enabling them across configurations (I)–(IV). As shown in Table 2, using semantic scanning alone (II) or tri-token prompting alone (III) leads to only marginal performance shifts relative to the baseline (I). However, when both components are activated together (IV), a distinct synergistic effect emerges, resulting in consistent gains across all evaluation metrics.

% \textbf{Effect of WKV-sharing/moment:}
% Transition from (IV) to (V) highlights significant role of WKV-sharing/moment mechanism. Building upon foundation of semantic scanning and tri-token prompting, addition of WKV-sharing/moment in (V) yields the most substantial performance improvement (e.g., PSNR: 42.3751, SSIM: 0.9737), confirming its pivotal function in integrating contextual information and refining model dynamics.

% \textbf{Effect of INN-shift:} The ablation sequence in Table 2 illustrates that the invertible neural network (INN)-shift operation is indispensable within the full architecture. The performance gap observed between configurations (I)–(IV) and the complete model (V) indicates that incorporating the INN-shift is essential for attaining peak performance. These results verify that its design—which guarantees lossless propagation of high-frequency information.

\begin{figure}[t]
	\centering
	\includegraphics[width=\columnwidth]{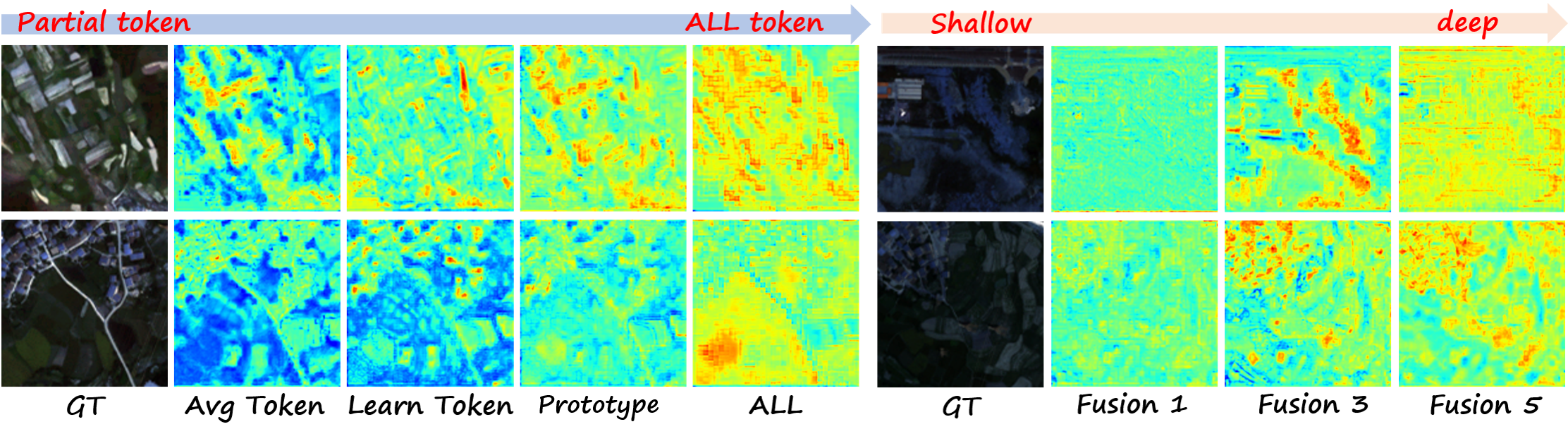}
	\caption{Feature analysis of prompt-learning tokens and model depth.}
	\label{fea}
\end{figure}

\textbf{Feature analysis:}  \cref{fea} demonstrates that the progressive integration of token processing strategies and model depth optimization is essential for achieving optimal feature representation.  The left portion reveals the evolution of token processing strategies, where `Avg Token' shows basic feature coverage, `Learn Token' demonstrates more focused activation patterns, and the full `ALL token' integration achieves comprehensive feature alignment with ground truth structures. Similarly, the right side illustrates depth-wise refinement, from elementary pattern  in shallow to semantically meaningful activations in deep layers.

% \clearpage % flush pending floats (e.g., figure*) before the conclusion
\section{Conclusion}
In this paper, we presented a novel Multigrain-aware Semantic Prototype Scanning paradigm for pan-sharpening. Our key contribution is a semantic-driven scanning strategy that replaces the recurrent positional bias of RWKV models with a clustering-based prototype guidance, enabling context-aware fusion. This is further enhanced by a tri-token prompting mechanism and an invertible Q-shift operation, collectively ensuring robust feature refinement and high-frequency detail preservation. Extensive experiments across multiple pan-sharpening benchmarks demonstrate that our approach achieves superior performance.

\newpage
\noindent\textbf{Acknowledgment.} This work has been supported in part by the National Natural Science Foundation of China (No. U24B20175, No. 62322216), Shenzhen Science and Technology Program (No. JCYJ20241202125904007, No. RCYX20221008092849068), the Foundation of Key Laboratory of Education Informatization for Nationalities (Yunnan Normal University), Ministry of Education (No. EIN2024B001), the Open Fund of Key Laboratory of the Ministry of Education on Artificial Intelligence in Equipment (No. 2024-AAIE-KF04-01).

%% file: sec/3_finalcopy.tex
% \section{Final copy}

% You must include your signed IEEE copyright release form when you submit your finished paper.
% We MUST have this form before your paper can be published in the proceedings.

% Please direct any questions to the production editor in charge of these proceedings at the IEEE Computer Society Press:
% \url{https://www.computer.org/about/contact}.